\documentclass[10pt,twocolumn,letterpaper]{article}

\usepackage{cvpr}              
\usepackage{booktabs}   
\usepackage{multirow}   
\usepackage{amssymb}    
\usepackage{graphicx}   
\usepackage{adjustbox}   
\usepackage{svg}      
\usepackage{subcaption}
\usepackage[table]{xcolor}
\usepackage{textcomp}
\usepackage{url}
\usepackage[accsupp]{axessibility} 

\svgpath{{images/}}

\definecolor{cvprblue}{rgb}{0.21,0.49,0.74}
\usepackage[pagebackref,breaklinks,colorlinks,allcolors=cvprblue]{hyperref}

\def\paperID{31335} 
\def\confName{CVPR}
\def\confYear{2026}

\title{SemLT3D: Semantic-Guided Expert Distillation for Camera-only Long-Tailed 3D Object Detection}

\author{
Hao Vo\textsuperscript{1},
Khoa Vo\textsuperscript{1},
Thinh Phan\textsuperscript{1},
Ngo Xuan Cuong\textsuperscript{1}, \\
Gianfranco Doretto\textsuperscript{2},
Hien Nguyen\textsuperscript{3},
Anh Nguyen\textsuperscript{4},
Ngan Le\textsuperscript{1} \\
\textsuperscript{1}AICV Lab, University of Arkansas, USA \quad
\textsuperscript{2}University of Utah, USA \\
\textsuperscript{3}University of Houston, USA \quad
\textsuperscript{4}University of Liverpool, UK \\[2pt]
{%
\ttfamily\small
\makebox[\linewidth][c]{%
\parbox{0.95\linewidth}{\centering
\tt\small \{haov,khoavoho,thinhp,cngo,thile\}@uark.edu, u6069230@umail.utah.edu \\
\tt\small hvnguy35@central.uh.edu, anh.nguyen@liverpool.ac.uk
}}%
}%
}

\begin{document}
\maketitle

\begin{abstract}

Camera-only 3D object detection has emerged as a cost-effective and scalable alternative to LiDAR for autonomous driving, yet existing methods primarily prioritize overall performance while overlooking the severe long-tail imbalance inherent in real-world datasets. In practice, many rare but safety-critical categories such as children, strollers, or emergency vehicles are heavily underrepresented, leading to biased learning and degraded performance. This challenge is further exacerbated by pronounced {inter-class ambiguity} (e.g., visually similar subclasses) and {substantial intra-class diversity} (e.g., objects varying widely in appearance, scale, pose, or context), which together hinder reliable long-tail recognition. In this work, we introduce \textbf{SemLT3D}, a Semantic-Guided Expert Distillation framework designed to enrich the representation space for underrepresented classes through semantic priors. SemLT3D consists of: (1) a \textbf{language-guided mixture-of-experts} module that routes 3D queries to specialized experts according to their semantic affinity, enabling the model to better disentangle confusing classes and specialize on tail distributions; and (2) a \textbf{semantic projection distillation} pipeline that aligns 3D queries with CLIP-informed 2D semantics, producing more coherent and discriminative features across diverse visual manifestations. Although motivated by long-tail imbalance, the semantically structured learning in SemLT3D also improves robustness under broader appearance variations and challenging corner cases, offering a principled step toward more reliable camera-only 3D perception.

\end{abstract}    
\section{Introduction}
\label{sec:intro}

\begin{figure}
    \centering
    \includegraphics[width=\linewidth]{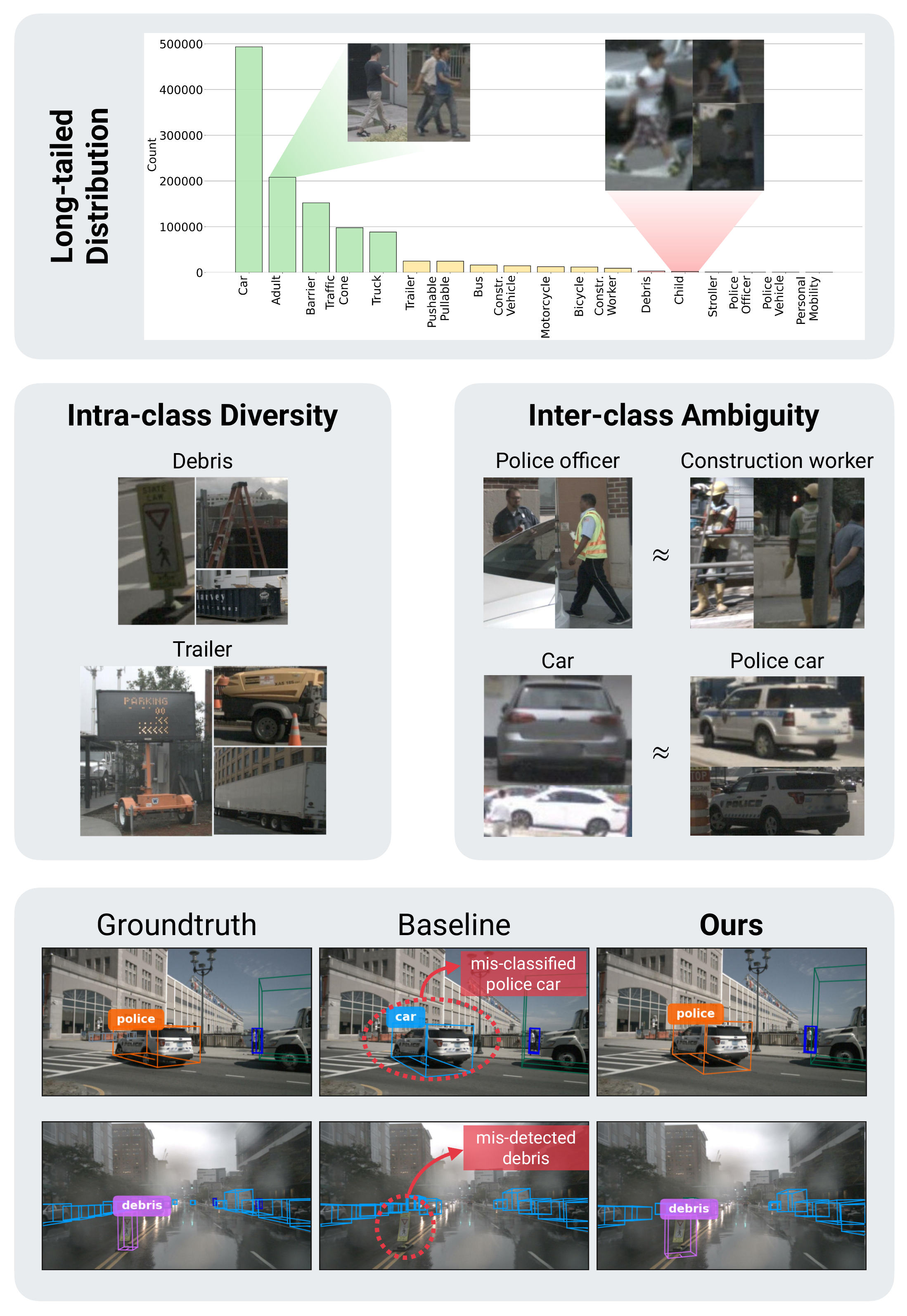}
\caption{Visualization of long-tailed distribution in nuScenes \cite{caesar2020nuscenes} (top), illustration of inter-intra categories diversity challenges in tail-classes (middle), and representative failure cases of baseline model caused by such inter class ambiguities (bottom-1$^{st}$) and intra-class diversity (bottom - 2$^{nd}$).}
\label{fig:intro_longtail}
    \vspace{-6mm}

\end{figure}

In recent years, camera-only 3D object detection for autonomous driving (AV) \cite{huang2021bevdet, wang2022detr3d, park2022time, huang2022bevdet4d, liu2022petr, liu2023sparsebev, wang2023exploring, li2024bevnext, jiang2024far3d, xue2025corrbev} has become one of the most active and transformative research topics in perception. Compared with LiDAR-based counterparts \cite{lang2019pointpillars, chen2023largekernel3d, liu2022bevfusion}, camera-only perception systems offer substantially lower hardware cost and easier scalability, enabling mass deployment at fleet level. A broad spectrum of methods have merged, ranging from depth-estimation–based approaches \cite{huang2021bevdet, li2023bevdepth} to depth-free paradigms \cite{li2024bevformer, liu2022petr, wang2022detr3d}, from single-frame detectors \cite{wang2022detr3d} to multi-frame temporal models \cite{liu2023sparsebev, wang2023exploring}, and from sparse-query \cite{liu2023sparsebev, lin2022sparse4d} to dense-query \cite{li2024bevformer, li2024bevnext} frameworks.

Despite these advances, existing camera-only 3D object detectors remain biased toward just frequently seen categories. Object occurrences in driving datasets follow a strongly long-tailed (Zipfian) distribution~\cite{zipf2013psycho}, where a few dominant \textit{``head"} classes such as \texttt{car} and \texttt{adult} appear abundantly, while many \textit{``tail"} classes (e.g., \texttt{child}, \texttt{emergency vehicle}, \texttt{stroller}, \texttt{debris}, etc.) are extremely underrepresented~\cite{zhang2023deep}. Yet these rare classes are often those with the highest safety stakes: a missed \texttt{child} or \texttt{emergency vehicle} detection can have catastrophic consequences~\cite{taeihagh2019governing, wong2020identifying}. In widely adopted benchmarks such as nuScenes~\cite{caesar2020nuscenes}, evaluation protocols further exacerbate this imbalance by aggregating distinct subcategories--\texttt{child}, \texttt{police officer}, \texttt{construction worker}, \texttt{stroller}--into a coarse group of pedestrian, obscuring crucial behavioral and safety nuances. As a result, current models achieve impressive overall performance yet remain unreliable in the rare but safety-critical cases that define real-world robustness.

Beyond data scarcity and imbalance, two inherent challenges further exacerbate the issue in camera-only 3D detection (Fig.~\ref{fig:intro_longtail} middle). First, intra-class diversity: tail categories often cover visually heterogeneous instances. For example, \texttt{debris} can refer to trash containers, ladders, or fallen cargo. Second, inter-class ambiguity: visual overlaps between semantically related classes cause frequent confusion. The \texttt{police officer} may wear reflective vests that are indistinguishable from the \texttt{construction worker}, and \texttt{the police car} at certain viewpoints may resemble a typical \texttt{car}.

To address these challenges, we introduce SemLT3D, a semantic-guided expert distillation for camera-only long-tailed 3D object detection. Following recent query-based paradigms for 3D detection \cite{wang2023exploring, jiang2024far3d}, SemLT3D initializes a set of 3D object queries that aggregate visual information from multiple camera views to form 3D object tokens, from which a detection head decodes into 3D object locations and categories. Building on this foundation, SemLT3D directly targets two major difficulties identified above through two complementary components. \textbf{(1) Language-guided mixture-of-experts} addresses intra-class diversity by introducing semantically guided expert specialization. The router leverages language embeddings of object names to guide the assignment of 3D object queries to experts according to their visual-language semantic similarity. This language-guided routing allows semantically related categories to share experts, while distinct categories are handled by different ones, enabling each expert to focus on a coherent and meaningful subset of visual patterns. \textbf{(2) Semantic projection distillation} mitigates data scarcity and inter-class ambiguity by transferring semantic knowledge from a powerful pretrained vision-language model (VLM) into our 3D object features. By aligning 3D object tokens with corresponding semantically grounded 2D image embeddings extracted from the pretrained VLM, we inject high-level semantic priors into 3D object tokens. This semantic transfer enriches the visual representations of underrepresented categories and improves discrimination among visually similar classes, leading to more robust long-tailed 3D detection.

To demonstrate the effectiveness and generality of our proposed method, we follow prior work \cite{peri2023towards} and extend the evaluation from the standard 10 classes to all 18 classes on both the nuScenes \cite{caesar2020nuscenes} and Argoverse 2 (AV2) \cite{wilson2023argoverse} datasets, providing a more comprehensive assessment of overall performance. Without any bells and whistles, our approach consistently improves upon the corresponding baselines by 2.62\%/ 0.96\% mAP and 2.75\%/ 1.62\% NDS on nuScenes with ResNet-50 and ResNet-101 backbones, respectively, and raises 1.5\% mAP and 1.3\% NDS on the AV2 validation split. In summary, our main contributions are as follows:
\begin{itemize}
    \item We propose \textbf{SemLT3D}, a semantic-guided expert distillation framework, that tackles long-tail imbalance, along with inter-class ambiguity, and intra-class diversity in camera-only 3D detection.
\item SemLT3D couples a \textbf{language-guided mixture-of-experts} which semantically routes queries for expert specialization with a \textbf{semantic projection distillation} that injects CLIP-informed 2D priors into 3D tokens for stronger class discrimination.
\item \textbf{Extensive evaluation} under the full 18-class long-tailed setting shows consistent gains across baselines, establishing SemLT3D as an effective and scalable solution for robust camera-only 3D perception.
\end{itemize}
\section{Related Work}

\textbf{Camera-based 3D Object Detection} Recently, camera-based 3D object detection has attracted much attention and achieved great progress,
due to its advantages of low deployment cost and rich semantic information. The related methods can be roughly categorized into two branches: BEV-based representation \cite{philion2020lift, huang2021bevdet, li2023bevdepth, li2023bevstereo} and query-based representation \cite{liu2023petrv2, wang2023exploring, li2024bevformer, yang2023bevformer, liu2023sparsebev}. While these methods have significantly advanced overall performance, handling corner cases may be the last mile of perception systems. Recent studies have thus focused on addressing specific challenges within camera-based 3D detection. For instance, RoboBEV \cite{xie2023robobev} adding more challenging attributes to the nuScenes benchmarks. Far3D \cite{jiang2024far3d} and LR3D \cite{yang2024improving} address the long-range problem by extend the detection range and using support from 2D detection or depth estimation. Motivated by human behavior, CorrBEV \cite{xue2025corrbev} resolve the occlusion by propose to use the correlation between text and image. \textit{In contrast, our work turns attention to a largely underexplored yet critical issue, the long-tail distribution in camera-based 3D object detection.}


\begin{figure*}[t]
    \centering
    \includegraphics[width=\textwidth]{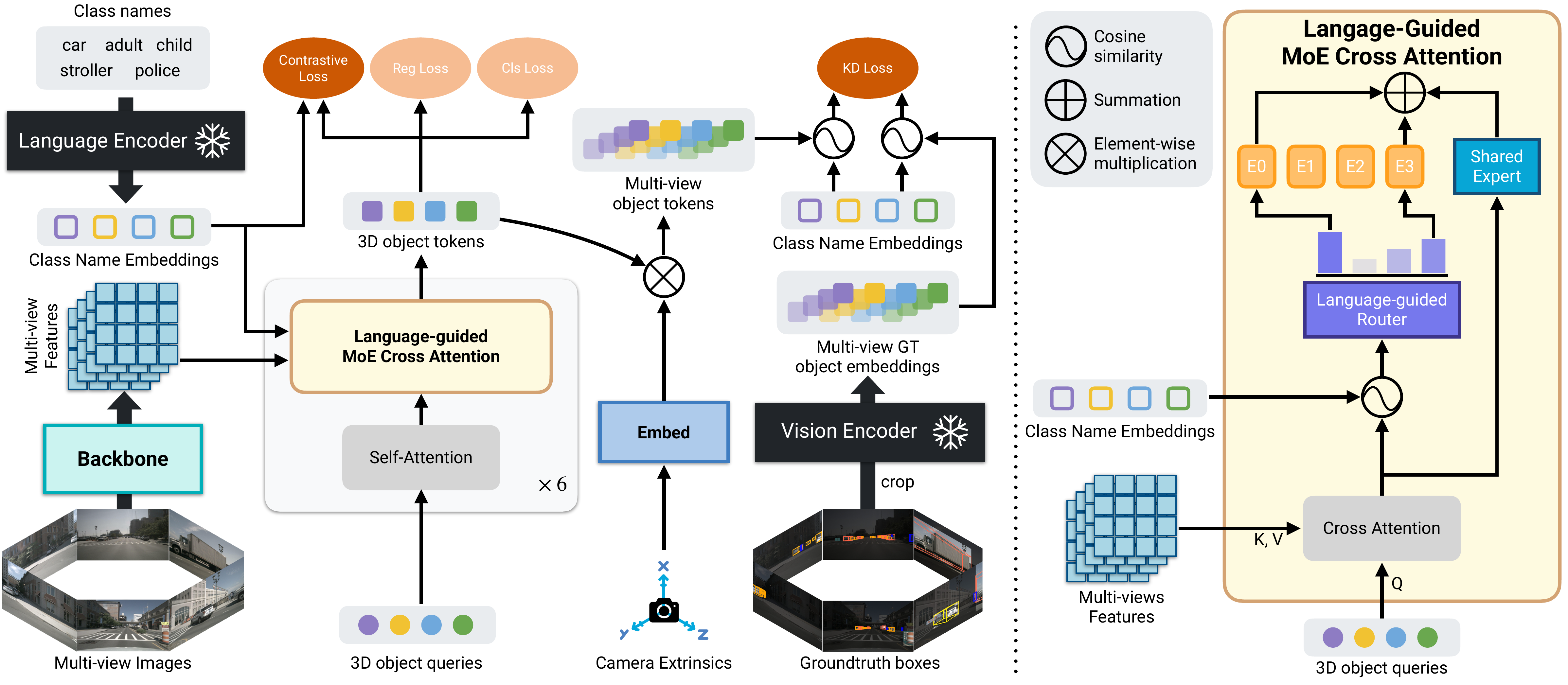}
    \caption{Overall architecture of the proposed SemLT3D for multi-view 3D long-tailed object detection.}
    \label{fig:overall_framework}

    \vspace{-4mm}
    
\end{figure*}

\textbf{3D Long-tail Object Detection} Resolving the long-tail problem remains an inevitable and critical challenge in autonomous driving, particularly for 3D object detection. A large body of prior work has focused on addressing this challenge through LiDAR-based pipelines. For instance, CBGS \cite{zhu2019class} mitigates data imbalance by up-sampling LiDAR sweeps of rare categories and pasting instances of underrepresented objects from other scenes, while LT3D \cite{peri2023towards} introduces a hierarchical loss that leverages superclass–subclass relations and camera–LiDAR fusion to improve recall on rare categories. More recently, FOMO-3D \cite{yangfomo} integrates OWLv2 \cite{minderer2023scaling} for 2D open-vocabulary detection and Metric3Dv2 \cite{hu2024metric3d} for depth estimation to enhance long-tail recognition. Although these approaches demonstrate notable progress, their reliance on LiDAR or multi-modal (LiDAR–camera) and multi-stage (e.g., LiDAR–VLM or LiDAR–2D detector) inference pipelines introduces additional computational overhead, synchronization complexity, hardware cost, and latency, making large-scale deployment less feasible. Surprisingly, despite these limitations, the long-tail issue in camera-only 3D detection has received little attention. \textit{Motivated by this observation, we propose \textbf{SemLT3D}, a simple yet effective plug-and-play framework that tackles the long-tail problem within a unified camera-only 3D detection paradigm, maintaining a favorable balance between scalability, efficiency, and practical deployment.}

\textbf{Connection to 2D Long-tail learning} To better understand imbalance in camera-based 3D detection, it is helpful to draw parallels with the extensive progress made in 2D long-tail learning~\cite{zhang2023deep}. In 2D classification, re-sampling~\cite{estabrooks2004multiple,kang2019decoupling} and re-weighting~\cite{lin2017focal,cui2019class,ren2020balanced} are widely used, yet they remain constrained by the inherently limited samples of tail classes and therefore cannot fully address representation deficiency. For 2D detection, prior work generally falls into two directions. First, multi-stage training pipelines inspired by few-shot detection~\cite{tran2025simltd,hu2020learning,meng2023learning,dong2023boosting} attempt to transfer head-class knowledge to tail classes; however, such approaches require training multiple models, multiple times and become prohibitively expensive for large-scale 3D benchmarks~\cite{wilson2023argoverse,caesar2020nuscenes}. Second, other methods leverage additional data or image-level supervision~\cite{meng2023learning,zhou2022detecting}, often using CLIP to enrich tail semantics, but these approaches face a significant domain gap when applied directly to 3D detection. \textit{Motivated by these insights, we introduce a language-guided MoE that routes queries by semantic similarity, allowing experts to share strong head-class features while dedicating capacity to tail categories, together with a semantic projection distillation mechanism that enables effective distillation from powerful 2D vision–language models. Taken together, these designs bring the benefits of mature 2D long-tail strategies into a unified and scalable camera-only 3D detection framework.}


\section{Method}
\label{sec:method}


We present SemLT3D (Figure~\ref{fig:overall_framework}) for camera-based 3D object detection. In section \ref{sec:cmoe}, we propose semantic-guided MoE by using language embeddings. Next, we delineate the process of semantic projection distillation in Section \ref{sec:2d_3d_kd} and process of visual-language alignment in  Section \ref{sec:query_language}.

\begin{figure}[t]
    \centering
    \includegraphics[width=0.9\linewidth]{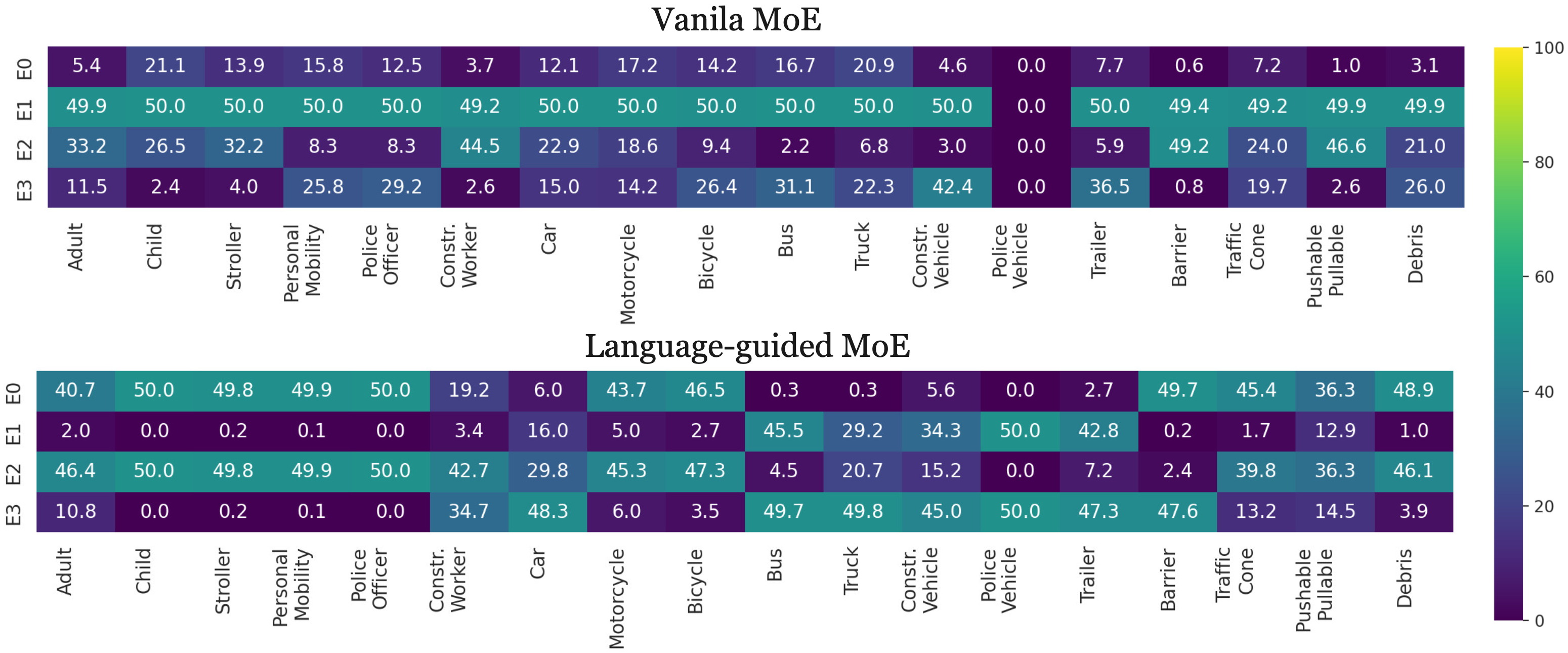}
\vspace{-2mm}
    
\caption{Distribution of queries assigned to activated experts for standard MoE and our LMoE. The x-axis shows category names and the y-axis indicates expert IDs.}
\label{fig:moe_routing}
    \vspace{-4mm}

\end{figure}
\subsection{Language-guided Mixture of Experts}
\label{sec:cmoe}
As aforementioned, learning from highly long-tailed diversity data with a unified model is inherently challenging, as optimization is dominated by head classes while tail classes receive insufficient supervision. This issue is further exacerbated by substantial intra-class diversity in rare categories - e.g., trailer, debris exhibit wide variations in pose, appearance, and context despite having very few training examples. To alleviate this limitation without increasing overall model complexity, we draw inspiration from recent advances in mixture-of-experts architectures \cite{dai2024deepseekmoe, lin2024moe} and replace the standard Feed-Forward Network (FFN) layers in transformer blocks with our proposed Language-guided Mixture-of-Experts (LMoE).


Unlike a conventional FFN that applies a single transformation to all object queries, LMoE decomposes the representation space into multiple lightweight experts, each specializing in a subset of semantically related categories. This category-level specialization reduces interference between heterogeneous classes and allows the model to better capture the unique appearance variations within tail categories. Furthermore, instead of treating every category uniformly, LMoE intentionally routes semantically related categories into the same expert based on their shared geometric and contextual characteristics. For example, human-like categories such as \texttt{adult}, \texttt{police officer} or \texttt{debris} tend to share similar height ranges, body proportions, and spatial placement in driving scenes, while vehicle-related categories exhibit their own consistent structural patterns. By assigning these semantically aligned categories to the same expert, LMoE forms meaningful sub-clusters that reduce intra-expert heterogeneity, making it easier to handle large intra-class diversity.

As illustrated in Figure~\ref{fig:overall_framework}, LMoE consists of a router, $M$ specialized experts, and one shared expert, all implemented as lightweight linear layers. For each query token, the router predicts routing weights and selects the top-$k$ experts to process it, after which their outputs are aggregated according to the routing weights. In parallel, each token is also passed through the shared expert to retain generalizable knowledge across all categories.

Recent DETR-style detectors, which directly adopt MoE layers designed for large-language-model architectures \cite{lu2025dynamic, zhang2025mr}, often promote uniform routing \cite{omi2025load}, limiting their ability to form meaningful semantic specialization. To address this shortcoming, we redesign the routing mechanism by replacing raw query features with language-aware similarity vectors. Instead of feeding high-dimensional query embeddings into the router, we compute their similarity to class name embeddings extracted by CLIP's text encoder, providing an explicit semantic prior that naturally groups queries associated with related meanings. As shown in Figure~\ref{fig:moe_routing}, this design produces clear semantic partitions, yielding expert assignments that align with intuitive structure—such as grouping human-like classes together or routing all vehicle-related categories into another expert. Concretely, after the cross-attention block, the 3D queries $Q \in \mathbb{R}^{k \times D}$ (where $k$ is the number of queries and $D$ is the query dimension) are projected into the language embedding space of dimension $d$, producing semantically aligned queries $\hat{Q} \in \mathbb{R}^{k \times d}$.
\begin{align}
    \hat{Q} &= \mathrm{Linear}(Q) 
\end{align}

\noindent then, we extract language-driven semantic embeddings by converting the category names into prompts \cite{zhong2022regionclip} and feeding into the CLIP language encoder, as result we obtain $P_{\text{language}} \in \mathbb{R}^{n \times d}$, where $n$ denotes the number of classes:
\vspace{-4mm}
\begin{align}
    P^{\text{language}}
    &= 
        \mathrm{CLIP}_{\text{language}}\!\left(\textit{categories}\right)
\end{align}

\noindent Next, we compute the similarity between $\hat{Q}$ and $P^{\text{language}}$ to obtain $S^l \in \mathbb{R}^{k \times n}$, which serves as the input to the router:
\vspace{-4mm}

\begin{align}
    S^l &= \text{sim}(\hat{Q} , P^\text{language})\\
    R &= Router(S^l)
\end{align}

\noindent where $\mathrm{sim}(\mathbf{a}, \mathbf{b}) = \frac{\mathbf{a}^\top \mathbf{b}}{\|\mathbf{a}\|\|\mathbf{b}\|}$. Based on the router output $R$, we apply a softmax function to obtain routing weights:

\begin{equation}
\begin{aligned}
    W &= \mathrm{Softmax}(R)
\end{aligned}
\end{equation}

\noindent These routing weights determine the contribution of each expert for every query. Specifically, each query is assigned to its top-$k$ experts according to $W$, and their outputs are aggregated to form the routed representation $y^{e} \in \mathbb{R}^{k \times D}$. In parallel, a shared expert $E^{s}$ processes all queries to retain generalizable knowledge. The final refined query representation $\bar{Q}$ is then computed as:
\begin{equation}
    y^{e} = \sum_{i \in \mathcal{T}} W_i E^{R}_{i}(Q), \qquad \bar{Q} = y^{e} + E^{s}(Q),
\end{equation}
\noindent where $\mathcal{T}$ denotes the set of top-$k$ selected expert indices and $E^{R}_{i}$ represents the $i$-th routed expert. Following prior work~\cite{dai2024deepseekmoe, bai2023qwen}, we further apply an auxiliary balance loss to encourage diverse expert utilization:
\begin{align}
    \mathcal{L}_{\text{balance}} = M \cdot \sum_{i=1}^{M} \mathcal{F}_i \cdot \mathcal{P}_i,
\end{align}
\noindent where $\mathcal{F}_i$ is the fraction of queries assigned to expert $E_i$, and $\mathcal{P}_i$ is the average routing probability for that expert.


%

\begin{table*}[t]
\centering
\caption{Performance comparison on validation split of nuScenes \cite{caesar2020nuscenes} with 18 categories. The best score is in \textbf{bold}. 
}
\vspace{-2mm}

\label{tab:main_val_set}
\tiny
\resizebox{0.8\textwidth}{!}{
\setlength{\tabcolsep}{3.5pt}
\begin{tabular}{c|lcccc@{\hspace{1.0\tabcolsep}}c@{\hspace{1.0\tabcolsep}}c@{\hspace{1.0\tabcolsep}}c@{\hspace{1.0\tabcolsep}}c} 
\toprule
\textbf{} & \textbf{Methods} &  \textbf{mAP}$\uparrow$  &\textbf{NDS}$\uparrow$  & \textbf{mATE}$\downarrow$ & \textbf{mASE}$\downarrow$   &\textbf{mAOE}$\downarrow$   &\textbf{mAVE}$\downarrow$   &\textbf{mAAE}$\downarrow$ &\textbf{FPS}$\uparrow$ \\
\midrule
\multirow{9}{*}{\rotatebox{90}{Resnet50}}

& BEVDet \cite{huang2021bevdet}  & 13.27 & 19.46 & 0.892 & 0.528 & 0.814 & 0.819 & 0.665 & 13.30\\
& BEVDet4D \cite{huang2022bevdet4d}   & 17.98 & 26.66 & 0.8517 & 0.4557 & 0.8597 & 0.4563 & 0.6095 & 13.18\\
& BEVFormer \cite{li2024bevformer}   & 15.51 & 23.25 & 0.9333 & \textbf{0.3200} & 1.00 & 0.6818 & 0.5151  & 18.18\\
& BEVDepth \cite{li2023bevdepth}   & 15.43 & 22.01 & 0.7939 & 0.4638 & 0.9624 & 0.7355 & 0.6145 & 10.30\\
& PETRv2 \cite{liu2023petrv2}   & 18.13 & 27.63 & 0.8806 & 0.3235 & 0.9013 & 0.5425 & 0.4951 & 14.92\\
& RayDN \cite{liu2024ray}  & 27.41 & 37.39 & 0.6856 & 0.3516 & 0.7612 & 0.3574 & \textbf{0.4762} & 31.25\\

& SparseBEV~\cite{liu2023sparsebev}   & 27.94 & 38.74 & \textbf{0.6459} & 0.3496 & 0.7190 & 0.2775 & 0.5312 & 20.8\\

& \cellcolor{gray!15}StreamPETR~\cite{wang2023exploring}  & \cellcolor{gray!15}26.97 & \cellcolor{gray!15}38.19 & \cellcolor{gray!15}0.6553 & \cellcolor{gray!15}0.3597 & \cellcolor{gray!15}0.6839 & \cellcolor{gray!15}0.2960 & \cellcolor{gray!15}0.5351 & \cellcolor{gray!15}\textbf{32.36}\\

& \cellcolor{gray!15}Ours & \cellcolor{gray!15}\textbf{29.59} & \cellcolor{gray!15}\textbf{40.94} & \cellcolor{gray!15}0.6938 & \cellcolor{gray!15}0.3335 & \cellcolor{gray!15}\textbf{0.6153} & \cellcolor{gray!15}\textbf{0.2524} & \cellcolor{gray!15}0.4909 & \cellcolor{gray!15}29.40\\

\midrule
\multirow{4}{*}{\rotatebox{90}{Resnet101}}
& 
BEVFormer~\cite{li2024bevformer}    & 24.65 & 37.05 & 0.7731 & 0.3228 & 0.6051 & 0.3458 & 0.4807 & 5.01\\
& SparseBEV~\cite{liu2023sparsebev}    & 29.78 & 34.87 & \textbf{0.6129} & 0.6082 & 1.6232 & 0.2796 & 0.5016 & 9.21\\

& \cellcolor{gray!15}StreamPETR~\cite{wang2023exploring}     & \cellcolor{gray!15}30.16 & \cellcolor{gray!15}41.17 & \cellcolor{gray!15}0.6863 & \cellcolor{gray!15}0.3209 & \cellcolor{gray!15}0.6339 & \cellcolor{gray!15}\textbf{0.2607} & \cellcolor{gray!15}0.4892  & \cellcolor{gray!15}\textbf{11.76}\\

& \cellcolor{gray!15}Ours &  \cellcolor{gray!15}\textbf{31.12} & \cellcolor{gray!15}\textbf{42.79} & \cellcolor{gray!15}0.6479 & \cellcolor{gray!15}\textbf{0.3019} & \cellcolor{gray!15}\textbf{0.5928} & \cellcolor{gray!15}0.2665 & \cellcolor{gray!15}\textbf{0.4677}  & \cellcolor{gray!15}10.80\\

\bottomrule
\end{tabular}}
\vspace{-4mm}

\end{table*}

\begin{table}
\centering
\caption{Comparisons on the Argoverse 2 \cite{Argoverse2} validation split. We evaluate 26 object categories with a range of 150 meters. $\ddagger$ shows the results take from Far3D \cite{jiang2024far3d}.}
\label{tab:av2_val}

\begin{adjustbox}{width=\linewidth} 
\begin{tabular}{lcccccc}
\toprule
\textbf{Methods} & \textbf{Backbone}  & \textbf{mAP}$\uparrow$ & \textbf{CDS}$\uparrow$ & \textbf{mATE}$\downarrow$ & \textbf{mASE}$\downarrow$ & \textbf{mAOE}$\downarrow$ \\
\midrule
BEVStereo~\cite{li2023bevstereo} $\ddagger$  & VoV-99   & 14.6 & 10.4 & 0.847 & 0.397 & 0.901 \\
SOLOFusion~\cite{park2022time} $\ddagger$ & VoV-99   & 14.9 & 10.6 & 0.934 & 0.425 & 0.779 \\
PETR~\cite{liu2022petr}$\ddagger$       & VoV-99   & 17.6 & 12.2 & 0.911 & 0.339 & 0.819 \\
Sparse4Dv2~\cite{lin2023sparse4d}$\ddagger$ & VoV-99   & 18.9 & 13.4 & 0.832 & 0.343 & 0.723 \\
StreamPETR~\cite{wang2023exploring} $\ddagger$ & VoV-99   & 20.3 & 14.6 & 0.843 & 0.321 & 0.650 \\
\rowcolor{gray!15}
Far3D~\cite{jiang2024far3d}$\ddagger$      & VoV-99   & 24.4 & 18.1 & 0.796 & \textbf{0.304} & 0.538 \\
\rowcolor{gray!15}
Ours       & VoV-99   & \textbf{25.9} & \textbf{19.4} & \textbf{0.792} & 0.308 & \textbf{0.480} \\
\bottomrule
\end{tabular}
\end{adjustbox}
\end{table}

\subsection{Semantic Projection Distillation}
\label{sec:2d_3d_kd}

Next, to mitigate inter-class ambiguity, we integrate CLIP \cite{radford2021learning} to distill large-scale vision–language priors into our 3D detector. Ambiguous categories, such as \texttt{police officer} versus \texttt{construction worker}, often rely on subtle contextual cues (e.g., standing on a street corner or near a construction site), which CLIP captures effectively. Motivated by this observation, our distillation module is designed with 3 objectives: (1) directly strengthen the feature learning of 3D object tokens from CLIP, (2) inject semantic priors that benefit rare or visually similar categories, and (3) promote class-wise disentanglement through language–visual alignment. To this end, we transform the refined 3D object tokens $\bar{Q}$ into 2D camera-aligned representations by embedding each camera’s extrinsic parameters, producing $Q_c \in \mathbb{R}^{C \times k \times d}$, where $C$ is the number of cameras. This alignment enables direct supervision of 3D object tokens using 2D CLIP features at multiple views, consistent with our first objective.
\begin{align}
    Q_c &= \mathrm{Linear}(\bar{Q}) \odot \mathrm{Linear}(E)
\end{align}




\noindent where $\odot$ denotes the Hadamard product, $E\in R^{C \times 16}$ denotes cameras's extrinsics. Next, after performing standard Hungarian matching~\cite{wang2022detr3d, liu2022petr} between ground-truth objects and queries, we project the 3D boxes of the matched ground-truth objects onto the 2D image planes to obtain the ground-truth 3D–2D correspondences for each query. Using these correspondences, we crop the perspective regions of the matched objects in each camera view and feed them through the CLIP image encoder to extract rich visual features that capture not only class semantics but also contextual cues from the surrounding regions (our second objective).
\begin{align}
    P^{\text{visual}}_g = \mathrm{CLIP}_{\text{visual}}(\tilde{I}_g)
\end{align}




\noindent where $\tilde{I}_g$ denotes the cropped image region g-th in the set of matched 2D ground-truth G. To align semantic across modalities, we compute cosine similarities between the camera-aligned queries $Q_c$ and the language embeddings, as well as between the extracted visual features and the language embeddings. These similarity distributions form the student–teacher pair used in our distillation process:
\begin{align}
    S^s_g &= \mathrm{sim}(Q^c_g,\, P^{\text{language}}) \\
    S^t_g &= \mathrm{sim}(P^{\text{visual}}_g,\, P^{\text{language}})
\end{align}
We supervise the model using a KL divergence loss, encouraging the student queries to match the teacher’s language–visual alignment (our third objective):
\begin{align}
    \mathcal{L}_{\mathrm{KD}}
    &= \frac{1}{G} \sum_{g=1}^{G}
       \mathcal{L}_{\mathrm{KL}}\!\left(S^s_g,\, S^t_g\right)
\end{align}
\subsection{Query-Language Alignment}
\label{sec:query_language}


In addition to the two modules described above, to stablize the training, we further encourage alignment between the queries and language embeddings by introducing a contrastive loss. Following prior works \cite{lin2024moe, liu2024grounding, li2022grounded}, we compute the dot-product similarity between each query and the language embeddings to obtain classification logits, and supervise these logits using a focal loss \cite{lin2017focal}:
\begin{align}
    \mathcal{L}_{\text{contrast}} 
        &= \mathcal{L}_{\text{Focal}}\!\left(\mathrm{sim}(\hat{Q},\, P_{\text{language}}),\, T\right)
\end{align}

\noindent where $T \in \mathbb{R}^{k \times n}$ denotes is the target matching result between queries and classes computed from the Hungarian matching. Consistent with DETR-style architectures, we apply this contrastive loss as an auxiliary loss at each decoder layer to facilitate stable and progressive alignment during optimization.


\section{Experiment}

\subsection{Datasets and Metrics}

We evaluate our approach on two large-scale autonomous driving benchmarks: nuScenes \cite{caesar2020nuscenes} and Argoverse 2 \cite{Argoverse2}.

\textbf{nuScenes} consists of 1000 driving scenes (approximately 20 seconds each), with 3D bounding box annotations on sampled keyframes. Following prior work \cite{liu2022bevfusion, yang2025fomo}, we adopt the extended 18-class taxonomy and additionally report mAP for three frequency groups: Many, Medium, and Few, to better characterize long-tail performance. For standard evaluation, we follow the official nuScenes metrics, reporting mean Average Precision (mAP) and the nuScenes Detection Score (NDS) across 18 categories.

\textbf{Argoverse 2} provides 1000 scenes of 15 seconds duration at 10\,Hz, split into 700/150/150 for train/val/test. It offers seven high-resolution ring cameras with full 360° coverage and includes 26 object categories within a 150\,m perception range. We follow the official protocol and report mAP, the Composite Detection Score (CDS), and true-positive metrics ATE, ASE, and AOE.

\subsection{Implementation Details}
On nuScenes benchmark \cite{caesar2020nuscenes}, we implement our model base on StreamPETR \cite{wang2023exploring}, trained with 60 epochs using Resnet50 and Resnet101 \cite{he2016deep}, for other configs, we follow \cite{wang2023exploring}. On AV2 benchmark \cite{wilson2023argoverse}, we implemented our model base on Far3D \cite{jiang2024far3d} with the same settings. We use the CLIP-B/16 for knowledge distillation. To maintain complexity, we set the hidden dimensions of shared expert as 1024 and each expert is 512 so with top k = 2 will we able to maintain the origin complexity of the original FFN which is set to 2048. Weights of contrastive loss, knowledge distillation loss and MoE balance loss is set to 1.0, 0.5 and 0.01, respectively.  For fair comparison, all the training experiments are performed with 8 $\times$ NVIDIA A100(40GB).

\begin{table}[t]
\centering
\caption{Long-tail break down mAP evaluation on validation split on nuScenes \cite{caesar2020nuscenes}. L denotes LiDAR. C denotes Camera. ${}^\dag$ shows the result take from FOMO-3D~\cite{yang2025fomo}. ${}^\ast$ shows method using ViT backbone~\cite{fang2024eva}. \textcolor{gray!100}{Methods shown in grey utilize multi-sensor inputs.}}

\label{tab:nuscenes_longtail}
\resizebox{\columnwidth}{!}{%
\begin{tabular}{lccccc}
\toprule
\textbf{Method} & \textbf{Modality} & \textbf{All} & \textbf{Many} & \textbf{Medium} & \textbf{Few} \\

\midrule
CenterPoint~\cite{yin2021center, peri2023towards} ${}^\dag$& L & 39.2 & 76.4 & 43.1 & 3.5 \\
CenterPoint~\cite{yin2021center, peri2023towards}${}^\dag$(w/ hier.) & L & 40.4 & 77.1 & 45.1 & 4.3 \\
BEVFusion-L~\cite{liu2022bevfusion}${}^\dag$ & L & 42.5 & 72.5 & 48.0 & 10.6 \\
\midrule
\color{gray!100}TransFusion~\cite{bai2022transfusion}${}^\dag$ & \color{gray!100}L+C & \color{gray!100}39.8 & \color{gray!100}73.9 & \color{gray!100}41.2 & \color{gray!100}9.8 \\
\color{gray!100}BEVFusion~\cite{liu2022bevfusion}${}^\dag$ & \color{gray!100}L+C & \color{gray!100}45.5 & \color{gray!100}75.5 & \color{gray!100}52.0 & \color{gray!100}12.8 \\
\color{gray!100}CMT~\cite{liu2022bevfusion}${}^\dag$ & \color{gray!100}L+C & \color{gray!100}44.4 & \color{gray!100}79.9 & \color{gray!100}53.0 & \color{gray!100}4.8 \\
\color{gray!100}MMF~\cite{peri2023towards}${}^\dag$ & \color{gray!100}L+C & \color{gray!100}43.6 & \color{gray!100}77.1 & \color{gray!100}49.0 & \color{gray!100}9.4 \\
\color{gray!100}MMLF~\cite{ma2023long}${}^\dag$ & \color{gray!100}L+C & \color{gray!100}51.4 & \color{gray!100}77.9 & \color{gray!100}59.4 & \color{gray!100}20.0 \\
\color{gray!100}FOMO-3D~\cite{yang2025fomo}${}^\dag$ & \color{gray!100}L+C & \color{gray!100}54.6 & \color{gray!100}79.9 & \color{gray!100}59.6 & \color{gray!100}27.6 \\

\midrule
BEVDet \cite{huang2021bevdet} & C & 13.27 & 27.7 & 14.31 & 0.0 \\
BEVFormer \cite{li2024bevformer}  & C & 15.51 & 34.6 & 14.91 & 0.3 \\
BEVDepth \cite{li2023bevdepth} & C & 15.43 & 36.46 & 13.51 & 0.13 \\
BEVDet4D \cite{huang2022bevdet4d} & C & 17.98 & 39.34 & 18.07 & 0.0 \\
PETRv2 \cite{liu2023petrv2} & C & 18.13 & 38.82 & 14.96 & 0.3 \\
RayDN \cite{liu2024ray} & C & 27.41 & 54.50 & 31.02 & 0.6 \\
SparseBEV~\cite{liu2023sparsebev}  & C & 27.94 & 54.18 & 29.67 & 4.1 \\

StreamPETR~\cite{wang2023exploring}  & C & 26.97 & 53.32 &	28.53 &	3.22 \\

\midrule
Ours  & C & 29.59 & 50.92 & 34.55 &  6.03\\
\textbf{Ours${}^\ast$} & C & \textbf{41.1} & \textbf{62.08} & \textbf{40.62} & \textbf{20.77} \\
\bottomrule

\end{tabular}}
\vspace{-4mm}

\end{table}

\begin{figure}[t]
    \centering
    \includegraphics[width=\linewidth]{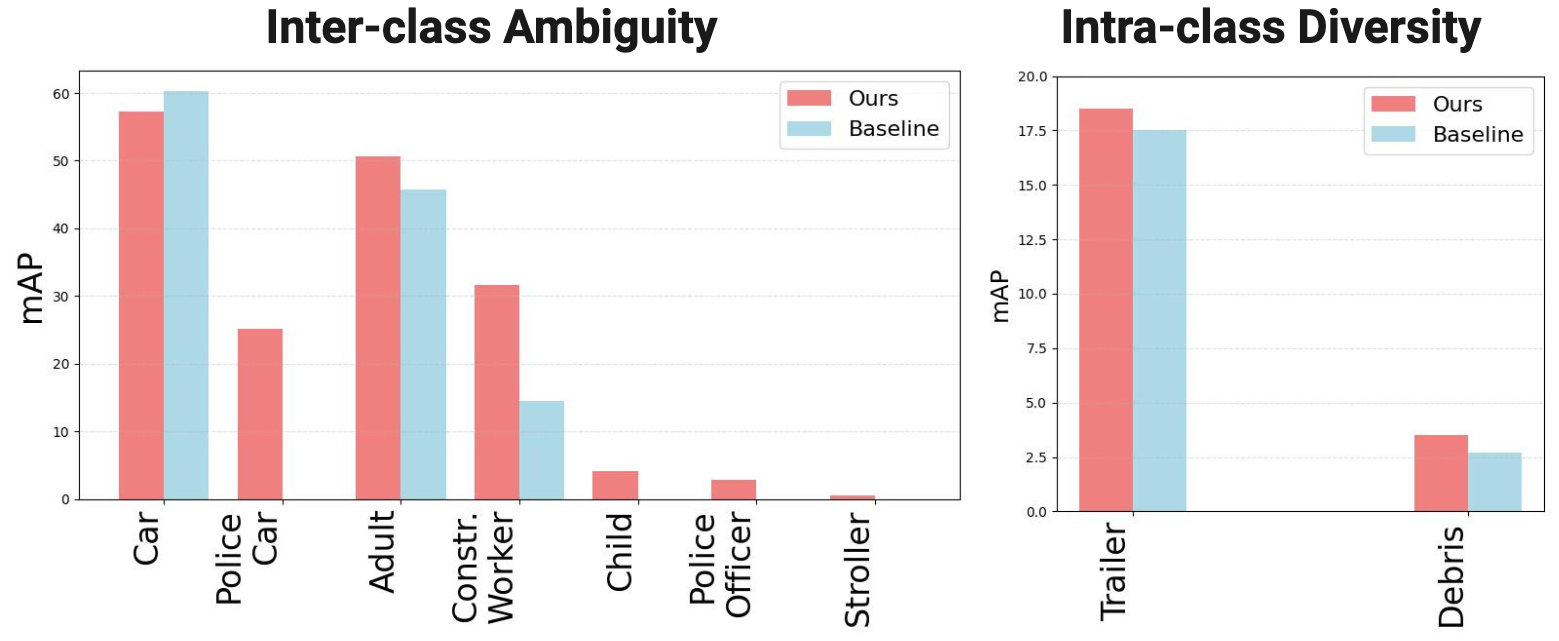}
    
    \caption{Performance comparison within tail categories has inter-intra diversity challenges.}
    \label{fig:inter-intra-quantitative}
    \vspace{-6mm}
    
\end{figure}


\subsection{State-of-the-art Comparison}

\textbf{Evaluation of Overall Performance.} To assess the effectiveness of SemLT3D on the nuScenes \cite{caesar2020nuscenes}, we re-evaluate eight camera-only baselines under the extended 18-class long-tailed setting, with results summarized in Table~\ref{tab:main_val_set}. Compared with the StreamPETR baseline \cite{wang2023exploring}, SemLT3D achieves substantial improvements of 2.62\% and 0.96\% mAP, as well as 2.75\% and 1.62\% NDS, on the ResNet50 and ResNet101 backbones, respectively. Importantly, our method introduces only a modest increase in inference time of 9.1\% for ResNet50 and 8.16\% for ResNet101, highlighting its efficiency. Similar trends are observed on the AV2 benchmark \cite{Argoverse2}, where SemLT3D improves Far3D \cite{jiang2024far3d} by 1.5\% mAP and 1.30\% CDS, as shown in Table~\ref{tab:av2_val}. These results demonstrate the strong effectiveness of our framework, along with its generalization ability and plug-and-play compatibility across datasets and model architectures, underscoring its practical deployment potential.

\begin{figure}[t]
    \centering
    \includegraphics[width=0.8\linewidth]{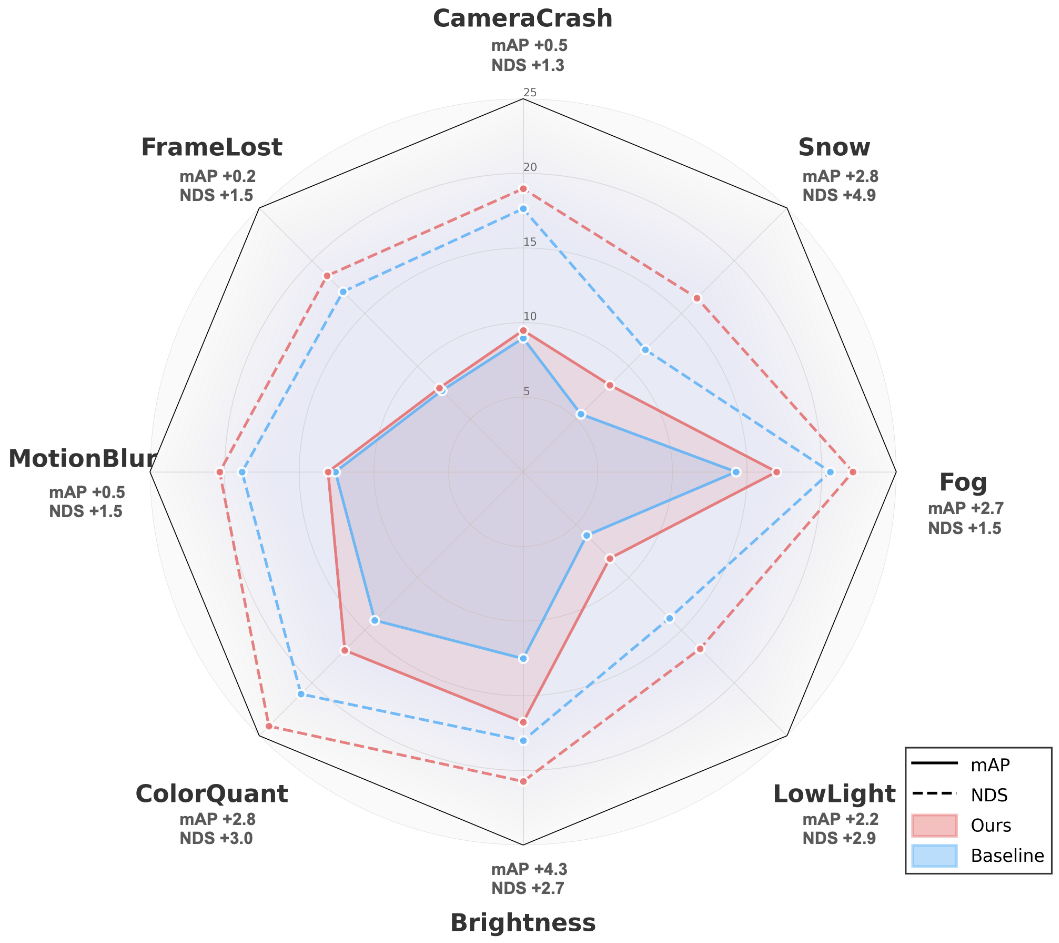}
\vspace{-2mm}
    
    \caption{Performance comparison under different corner cases.}
    \label{fig:corner-cases}
    \vspace{-6mm}
\end{figure}

\noindent \textbf{Evaluation of Long-tailed Performance.}  Beyond improving overall detection accuracy, our framework substantially enhances performance on classes with limited training samples. As shown in Table \ref{tab:nuscenes_longtail}, the proposed method yields notable gains of +6.02 mAP on Medium categories and +2.81 mAP on Few categories, with only a marginal decrease on Many. This aligns with our objective: by incorporating semantic-guided MoE and semantic projection distillation, the model leverages richer representations and strong semantic priors to better generalize to tail classes despite their scarcity. To further compare with multimodal approaches, we evaluate our framework using a ViT backbone \cite{fang2024eva}. The results demonstrate competitive performance relative to CenterPoint (+0.7 mAP) and BEVFusion (-1.3 mAP). Remarkably, even without relying on high-cost LiDAR sensors, our camera-only method achieves a +10.17 mAP improvement over BEVFusion on Few categories, illustrating the scalability and practical value of our approach for long-tailed detection.

\noindent \textbf{Evaluation on Inter–Intra Diverse Categories.} Figure~\ref{fig:inter-intra-quantitative} compares our SemLT3D with the baseline StreamPETR~\cite{wang2023exploring}. We consistently see better performance in both inter-class ambiguity cases and intra-class diversity cases, showing that our method is aligned with its objective, help the model distinguish confusing classes and handle the wide variations within each category. There is a small drop in the \texttt{car} class, likely due to overall optimization trade-offs, but it does not affect the general trend. Overall, the results highlight that SemLT3D brings clear and meaningful gains, especially for tail classes where diversity and ambiguity pose the biggest challenges.

\noindent \textbf{Evaluation on Other Corner Cases.} As discussed above, our motivation arises from the inherent diversity and ambiguity among categories. To further assess the robustness of our approach, we evaluate performance under challenging scenarios \cite{xue2025corrbev, xie2023robobev} where object appearances exhibit higher variability, such as \texttt{Snow} or \texttt{ColorQuant}. As illustrated in Figure~\ref{fig:corner-cases}, integrating our model into the baseline \cite{wang2023exploring} consistently improves detection performance across all scenarios, particularly where visual degradation or domain shifts weaken object representations. These results demonstrate that our model enhances the baseline’s generalization ability, enabling more reliable recognition under diverse and adverse conditions.


\begin{table}[t]
  \centering
  \caption{Effectiveness of each module on nuScenes val split.}
\vspace{-2mm}

  \label{tab:ablation_modules}
  \resizebox{\linewidth}{!}{
  \begin{tabular}{cccc|cc|ccc}
    \toprule
    \textbf{\shortstack{Mixture\\of\\Expert}} & \textbf{\shortstack{Semantic-\\Guided\\Router}} & \textbf{\shortstack{Semantic\\Projection\\Distillation}} & \textbf{\shortstack{Query-\\Language\\Alignment}}  & \textbf{mAP} & \textbf{NDS} & \textbf{Many} & \textbf{Medium} & \textbf{Few}   \\
    \midrule
    --          & --  & -- & --               & 26.97 & 38.19  & 53.32 &	28.53 &	3.22  \\
    --          &  -- & \checkmark   & --      & 28.30 & 39.13  & 49.6 & 32.1 & 5.82 \\
    \checkmark  & --  & --         & --     &  26.37 & 37.47  & 51.48 & 30.73 & 0.32 \\
    \checkmark  & \checkmark  & -- & --              & 27.50 &  37.59 & 51.7 & 32.16	& 2.25 \\
    \checkmark  & \checkmark   & \checkmark  & --           & 28.56 & 40.26 & 49.48	& 32.2 & 6.9 \\
    \checkmark  & \checkmark   & \checkmark  & \checkmark           & \textbf{29.59} & \textbf{40.94} & \textbf{50.92} &	\textbf{34.56} &	\textbf{6.03} \\
    
    \bottomrule
  \end{tabular}
  }
\vspace{-2mm}
  
\end{table}

\begin{table}[t]
  \centering
  \caption{Configurations of LMoE ablation on nuScenes val split.}
\vspace{-2mm}
  
  \label{tab:ablation_moe}
  \resizebox{0.8\linewidth}{!}{
  \begin{tabular}{cc|cc|ccc}
    \toprule
    \textbf{Experts} & \textbf{Top-\textit{k}}  & \textbf{mAP} & \textbf{NDS} & \textbf{Many} & \textbf{Medium} & \textbf{Few}   \\
    \midrule
    4          & 1         &  28.56 & 40.26 & 49.50 & 33.20  & 5.90 \\
    4          & 2         &  \textbf{29.59} & \textbf{40.94} & \textbf{50.92} & \textbf{34.55}  & 6.03 \\
    8          & 1         &   28.24 & 40.4 & 47.84 &  31.76 & 8.6 \\
    8          & 2         &  28.54 & 39.24 & 47.96 &  32.06 & \textbf{8.21} \\
    
    \bottomrule
  \end{tabular}
  }
\vspace{-2mm}
  
\end{table}

\begin{table}[t]
  \centering
  \caption{Ablation of prior semantic provided by
different CLIP models on nuScenes validation split.}
\vspace{-2mm}

  \label{tab:ablation_clip_models}
  \resizebox{0.8\linewidth}{!}{
  \begin{tabular}{c|cc|ccc}
    \toprule
    \textbf{CLIP models}  & \textbf{mAP} & \textbf{NDS} & \textbf{Many} & \textbf{Medium} & \textbf{Few}   \\
    \midrule
    ViT-B/32                   &  28.29 & 39.13 & 50.5 & 32.30  & 5.82 \\
    ViT-B/16                   &  29.59 & 40.94 & 50.92 & 34.55  & 6.03 \\
    ViT-L/14                   &  29.63 & 40.67 & 49.50 & 33.90  & 8.06 \\
    \bottomrule
  \end{tabular}
  }
    \vspace{-5mm}
  
\end{table}

\begin{figure*}[t]
    \centering
    \includegraphics[width=0.9\linewidth]{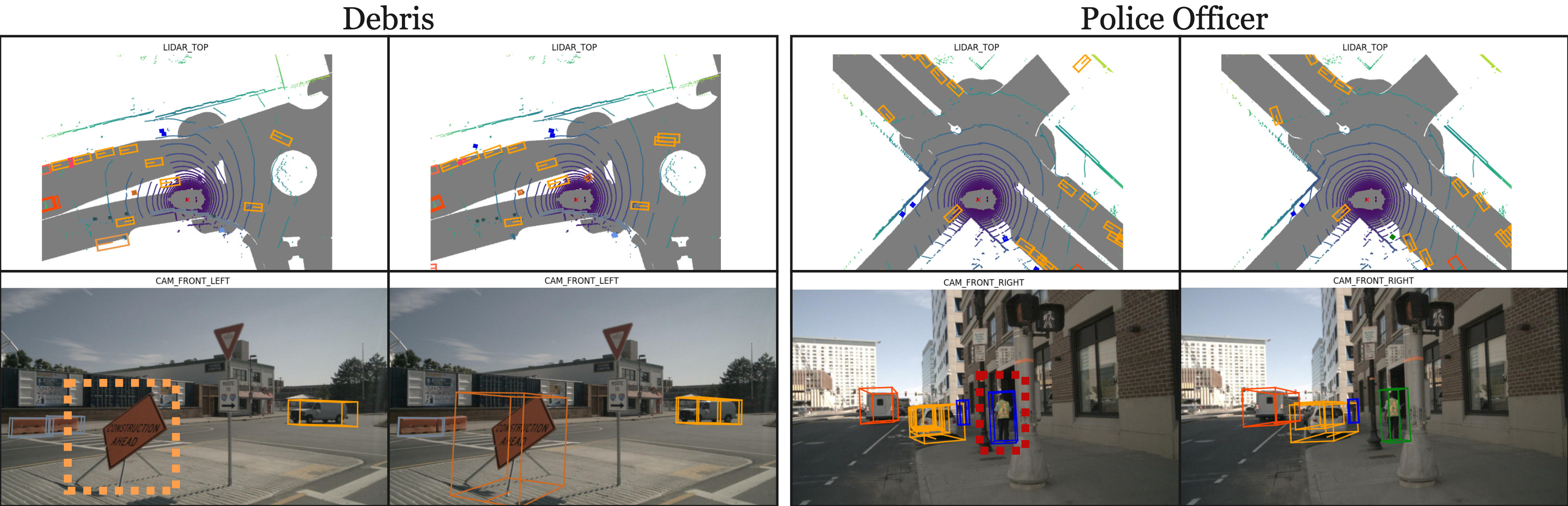}
\vspace{-2mm}
    
\caption{Quanlitative result on inter-intra diversity cases (Debris and Police Officer), compare between our method and baseline \cite{wang2023exploring} on tail-class. Left: our basline, Right: our proposed SemLT3D. The \textcolor{orange}{orange} dashed box highlight failure cases caused by intra-class diversity, while the \textcolor{red}{red} dashed box indicate failure cases arising from inter-class ambiguity.}

\label{fig:quanlitative}
    \vspace{-4mm}

\end{figure*}





\subsection{Ablation Study}

\textbf{Component-wise Analysis} Table~\ref{tab:ablation_modules} summarizes the effectiveness of each module and their impact on long-tail performance. Replacing the FFN with a vanilla MoE slightly decreases overall mAP (–0.6) and severely harms tail categories due to unguided expert routing. Introducing our semantic-guided router reverses this effect, improving both overall accuracy and rare-class performance by stabilizing expert assignment under long-tail distributions. CLIP-based semantic projection distillation further boosts performance from 26.97 to 28.30 mAP, particularly benefiting medium and tail classes that depend on strong contextual semantics. When combined with semantic-guided MoE, the synergy improves mAP to 28.56. Adding the query–language contrastive loss yields the best results, 29.59 mAP and 40.94 NDS, demonstrating that semantic routing, CLIP priors, and language-aligned supervision together produce more discriminative and balanced representations across Many/Medium/Few splits.

\noindent  \textbf{Semantic-guided MoE Configuration Analysis.} Table~\ref{tab:ablation_moe} evaluates different configurations of our LMoE module. Increasing the number of experts consistently improves performance on tail categories, suggesting that a larger expert pool captures richer and more fine-grained semantics crucial for rare classes. However, excessively expanding the expert set or reducing the top-\textit{k} selection leads to imbalanced routing and under-utilized experts, which negatively affects overall performance, especially on Many and Medium categories. Balancing specialization and stable expert usage, the configuration with 4 experts and top-\textit{k}=2 achieves the best trade-off, yielding the strongest results across both head and tail splits.


\begin{figure}[t]
    \centering
    \includegraphics[width=\linewidth]{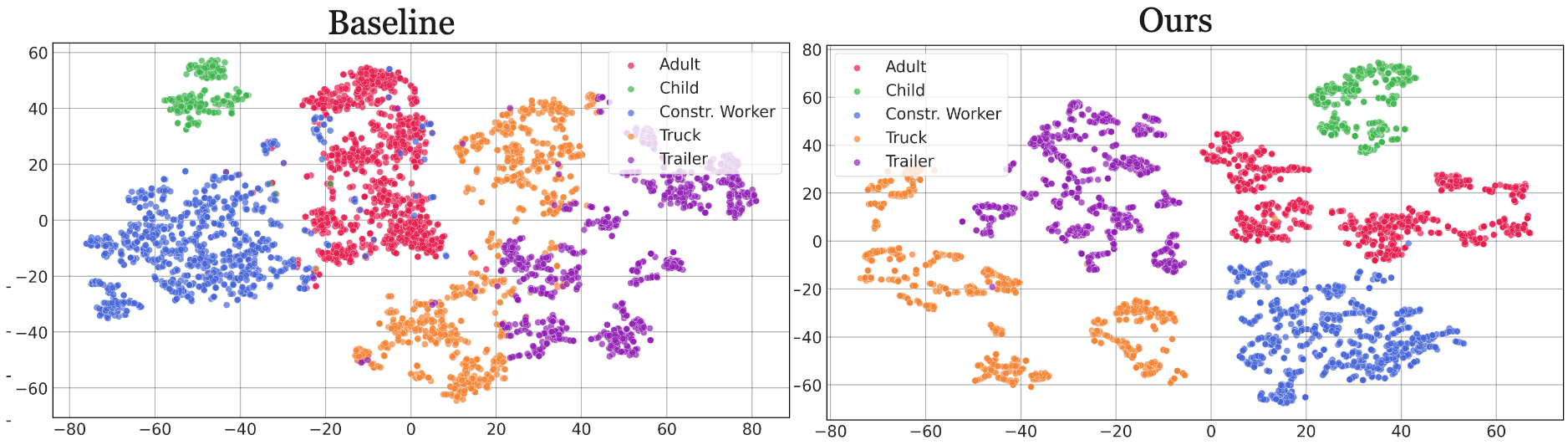}
    \caption{Inter-Intra class diversity visualization in embedding space via t-SNE. Left: our baseline \cite{wang2023exploring}, Right: our SemLT3D.}
    \label{fig:tnse_distritubtion}
    \vspace{-6mm}
    
\end{figure}

\noindent \textbf{Different CLIP Model Analysis.} Table~\ref{tab:ablation_clip_models} presents an ablation study on different variants of CLIP models. Although using a larger backbone such as ViT-L/14 yields the highest performance, it also significantly increases both training and inference time. To balance efficiency and accuracy, we adopt the ViT-B/16 variant as the teacher model in our final framework, which provides strong performance with moderate computational cost.

\subsection{Further Analysis}


\textbf{Qualitative Results} Figure \ref{fig:quanlitative} provides a qualitative comparison between the baseline StreamPETR and our proposed SemLT3D. Our focus is on addressing the persistent failures on tail classes and mitigating the challenges posed by both intra-class diversity and inter-class similarity. The figure highlights two representative failure cases where SemLT3D delivers clear improvements. First, SemLT3D successfully localizes the “debris,” whereas the baseline either misses it entirely (\textcolor{orange}{orange} box), illustrating how scarce samples and strong intra-class diversity can lead to confusion. Second, for the inter-class ambiguity distinction between “police officer” and “adult” SemLT3D correctly identifies the police officer while the baseline misclassifies it as an adult (\textcolor{red}{red} box), reflecting improved robustness against inter-class ambiguity. Overall, these qualitative results demonstrate the effectiveness of SemLT3D in handling tail classes and improving reliability in safety-critical scenarios for autonomous driving.

\noindent\textbf{Inter–Intra Class Distribution.} Figure~\ref{fig:tnse_distritubtion} compares the feature distributions of StreamPETR~\cite{wang2023exploring} and our SemLT3D, illustrating how our method reduces both inter-class ambiguity and intra-class variability—two challenges that are especially pronounced in tail categories. Using t-SNE~\cite{JMLR:v23:21-0524}, we visualize the features of three commonly confused classes (\texttt{adult}, \texttt{child}, \texttt{construction worker}) and observe that SemLT3D forms more compact and clearly separated clusters, indicating that our 2D–3D semantic distillation and query–language contrastive alignment effectively mitigate inter-class ambiguity. In parallel, the LMoE module enhances modeling of intra-class diversity by routing queries to specialized experts, enabling SemLT3D to produce distinct sub-clusters for different “trailer’’ variants, whereas the baseline remains scattered. These results demonstrate that our semantic alignment and expert specialization strategies substantially strengthen feature discrimination and reduce tail-class confusion.

\vspace{-4mm}

\section{Conclusion}
\vspace{-2mm}

In this paper, we introduce SemLT3D, a plug-and-play framework designed to address the long-tail challenge in camera-based multi-view 3D object detection. The central idea is to enhance representation learning for rare and visually ambiguous categories by using semantic priors through language-guided expert routing and 3D–2D knowledge distillation. With these components, SemLT3D delivers consistent improvements across multiple baseline detectors. We also provide a comprehensive evaluation under long-tail distributions and various corner-case scenarios, offering practical insights into the behavior of camera-only 3D perception systems. We believe that our semantic-guided design aligns with real deployment needs in autonomous driving and encourages further attention toward improving long-tail robustness in camera-based 3D detection.

\noindent \section*{Acknowledgement} 
\vspace{-2mm}

This material is based upon work supported by the National Science Foundation under Award No. \#2443877 (NSF CAREER), \#1946391 (NSF RII Track-1 DART), \#2445877 (NSF E-RISE). We also acknowledge the Arkansas High-Performance Computing Center for providing GPUs.

{
    \small
    \bibliographystyle{ieeenat_fullname}
    \bibliography{main}

@String(AAAI = {AAAI})

@article{huang2021bevdet,
  title={Bevdet: High-performance multi-camera 3d object detection in bird-eye-view},
  author={Huang, Junjie and Huang, Guan and Zhu, Zheng and Ye, Yun and Du, Dalong},
  journal={arXiv preprint arXiv:2112.11790},
  year={2021}
}

@inproceedings{wang2022detr3d,
  title={Detr3d: 3d object detection from multi-view images via 3d-to-2d queries},
  author={Wang, Yue and Guizilini, Vitor Campagnolo and Zhang, Tianyuan and Wang, Yilun and Zhao, Hang and Solomon, Justin},
  booktitle={Conference on robot learning},
  pages={180--191},
  year={2022},
  organization={PMLR}
}

@inproceedings{liu2022petr,
  title={Petr: Position embedding transformation for multi-view 3d object detection},
  author={Liu, Yingfei and Wang, Tiancai and Zhang, Xiangyu and Sun, Jian},
  booktitle={European conference on computer vision},
  pages={531--548},
  year={2022},
  organization={Springer}
}

@inproceedings{lang2019pointpillars,
  title={Pointpillars: Fast encoders for object detection from point clouds},
  author={Lang, Alex H and Vora, Sourabh and Caesar, Holger and Zhou, Lubing and Yang, Jiong and Beijbom, Oscar},
  booktitle={Proceedings of the IEEE/CVF conference on computer vision and pattern recognition},
  pages={12697--12705},
  year={2019}
}

@inproceedings{liu2023sparsebev,
  title={Sparsebev: High-performance sparse 3d object detection from multi-camera videos},
  author={Liu, Haisong and Teng, Yao and Lu, Tao and Wang, Haiguang and Wang, Limin},
  booktitle={Proceedings of the IEEE/CVF international conference on computer vision},
  pages={18580--18590},
  year={2023}
}

@inproceedings{chen2023largekernel3d,
  title={Largekernel3d: Scaling up kernels in 3d sparse cnns},
  author={Chen, Yukang and Liu, Jianhui and Zhang, Xiangyu and Qi, Xiaojuan and Jia, Jiaya},
  booktitle={Proceedings of the IEEE/CVF conference on computer vision and pattern recognition},
  pages={13488--13498},
  year={2023}
}

@inproceedings{xue2025corrbev,
  title={CorrBEV: Multi-View 3D Object Detection by Correlation Learning with Multi-modal Prototypes},
  author={Xue, Ziteng and Guo, Mingzhe and Fan, Heng and Zhang, Shihui and Zhang, Zhipeng},
  booktitle={Proceedings of the Computer Vision and Pattern Recognition Conference},
  pages={27413--27423},
  year={2025}
}

@article{huang2022bevdet4d,
  title={Bevdet4d: Exploit temporal cues in multi-camera 3d object detection},
  author={Huang, Junjie and Huang, Guan},
  journal={arXiv preprint arXiv:2203.17054},
  year={2022}
}

@inproceedings{li2023bevdepth,
  title={Bevdepth: Acquisition of reliable depth for multi-view 3d object detection},
  author={Li, Yinhao and Ge, Zheng and Yu, Guanyi and Yang, Jinrong and Wang, Zengran and Shi, Yukang and Sun, Jianjian and Li, Zeming},
  booktitle={Proceedings of the AAAI conference on artificial intelligence},
  volume={37},
  number={2},
  pages={1477--1485},
  year={2023}
}

@inproceedings{jiang2024far3d,
  title={Far3d: Expanding the horizon for surround-view 3d object detection},
  author={Jiang, Xiaohui and Li, Shuailin and Liu, Yingfei and Wang, Shihao and Jia, Fan and Wang, Tiancai and Han, Lijin and Zhang, Xiangyu},
  booktitle={Proceedings of the AAAI conference on artificial intelligence},
  volume={38},
  number={3},
  pages={2561--2569},
  year={2024}
}

@inproceedings{peri2023towards,
  title={Towards long-tailed 3d detection},
  author={Peri, Neehar and Dave, Achal and Ramanan, Deva and Kong, Shu},
  booktitle={Conference on Robot Learning},
  pages={1904--1915},
  year={2023},
  organization={PMLR}
}

@article{lin2022sparse4d,
  title={Sparse4d: Multi-view 3d object detection with sparse spatial-temporal fusion},
  author={Lin, Xuewu and Lin, Tianwei and Pei, Zixiang and Huang, Lichao and Su, Zhizhong},
  journal={arXiv preprint arXiv:2211.10581},
  year={2022}
}

@inproceedings{li2024bevnext,
  title={Bevnext: Reviving dense bev frameworks for 3d object detection},
  author={Li, Zhenxin and Lan, Shiyi and Alvarez, Jose M and Wu, Zuxuan},
  booktitle={Proceedings of the IEEE/CVF conference on computer vision and pattern recognition},
  pages={20113--20123},
  year={2024}
}

@article{zhu2019class,
  title={Class-balanced grouping and sampling for point cloud 3d object detection},
  author={Zhu, Benjin and Jiang, Zhengkai and Zhou, Xiangxin and Li, Zeming and Yu, Gang},
  journal={arXiv preprint arXiv:1908.09492},
  year={2019}
}

@article{taeihagh2019governing,
  title={Governing autonomous vehicles: emerging responses for safety, liability, privacy, cybersecurity, and industry risks},
  author={Taeihagh, Araz and Lim, Hazel Si Min},
  journal={Transport reviews},
  volume={39},
  number={1},
  pages={103--128},
  year={2019},
  publisher={Taylor \& Francis}
}

@inproceedings{wong2020identifying,
  title={Identifying unknown instances for autonomous driving},
  author={Wong, Kelvin and Wang, Shenlong and Ren, Mengye and Liang, Ming and Urtasun, Raquel},
  booktitle={Conference on Robot Learning},
  pages={384--393},
  year={2020},
  organization={PMLR}
}

@inproceedings{caesar2020nuscenes,
  title={nuscenes: A multimodal dataset for autonomous driving},
  author={Caesar, Holger and Bankiti, Varun and Lang, Alex H and Vora, Sourabh and Liong, Venice Erin and Xu, Qiang and Krishnan, Anush and Pan, Yu and Baldan, Giancarlo and Beijbom, Oscar},
  booktitle={Proceedings of the IEEE/CVF conference on computer vision and pattern recognition},
  pages={11621--11631},
  year={2020}
}

@article{dai2024deepseekmoe,
  title={Deepseekmoe: Towards ultimate expert specialization in mixture-of-experts language models},
  author={Dai, Damai and Deng, Chengqi and Zhao, Chenggang and Xu, RX and Gao, Huazuo and Chen, Deli and Li, Jiashi and Zeng, Wangding and Yu, Xingkai and Wu, Yu and others},
  journal={arXiv preprint arXiv:2401.06066},
  year={2024}
}

@inproceedings{tran2025simltd,
  title={SimLTD: Simple Supervised and Semi-Supervised Long-Tailed Object Detection},
  author={Tran, Phi Vu},
  booktitle={Proceedings of the Computer Vision and Pattern Recognition Conference},
  pages={4672--4681},
  year={2025}
}

@inproceedings{hu2020learning,
  title={Learning to segment the tail},
  author={Hu, Xinting and Jiang, Yi and Tang, Kaihua and Chen, Jingyuan and Miao, Chunyan and Zhang, Hanwang},
  booktitle={Proceedings of the IEEE/CVF conference on computer vision and pattern recognition},
  pages={14045--14054},
  year={2020}
}

@article{wilson2023argoverse,
  title={Argoverse 2: Next generation datasets for self-driving perception and forecasting},
  author={Wilson, Benjamin and Qi, William and Agarwal, Tanmay and Lambert, John and Singh, Jagjeet and Khandelwal, Siddhesh and Pan, Bowen and Kumar, Ratnesh and Hartnett, Andrew and Pontes, Jhony Kaesemodel and others},
  journal={arXiv preprint arXiv:2301.00493},
  year={2023}
}

@book{zipf2013psycho,
  title={The psycho-biology of language: An introduction to dynamic philology},
  author={Zipf, George Kingsley},
  year={2013},
  publisher={Routledge}
}

@article{meng2023learning,
  title={Learning from rich semantics and coarse locations for long-tailed object detection},
  author={Meng, Lingchen and Dai, Xiyang and Yang, Jianwei and Chen, Dongdong and Chen, Yinpeng and Liu, Mengchen and Chen, Yi-Ling and Wu, Zuxuan and Yuan, Lu and Jiang, Yu-Gang},
  journal={Advances in Neural Information Processing Systems},
  volume={36},
  pages={78082--78094},
  year={2023}
}

@inproceedings{yangfomo,
  title={FOMO-3D: Using Vision Foundation Models for Long-Tailed 3D Object Detection},
  author={Yang, Anqi Joyce and Tu, James and Dvornik, Nikita and Li, Enxu and Urtasun, Raquel},
  booktitle={9th Annual Conference on Robot Learning}
}

@inproceedings{dong2023boosting,
  title={Boosting long-tailed object detection via step-wise learning on smooth-tail data},
  author={Dong, Na and Zhang, Yongqiang and Ding, Mingli and Lee, Gim Hee},
  booktitle={Proceedings of the IEEE/CVF International Conference on Computer Vision},
  pages={6940--6949},
  year={2023}
}

@article{zhang2023deep,
  title={Deep long-tailed learning: A survey},
  author={Zhang, Yifan and Kang, Bingyi and Hooi, Bryan and Yan, Shuicheng and Feng, Jiashi},
  journal={IEEE transactions on pattern analysis and machine intelligence},
  volume={45},
  number={9},
  pages={10795--10816},
  year={2023},
  publisher={IEEE}
}

@inproceedings{radford2021learning,
  title={Learning transferable visual models from natural language supervision},
  author={Radford, Alec and Kim, Jong Wook and Hallacy, Chris and Ramesh, Aditya and Goh, Gabriel and Agarwal, Sandhini and Sastry, Girish and Askell, Amanda and Mishkin, Pamela and Clark, Jack and others},
  booktitle={International conference on machine learning},
  pages={8748--8763},
  year={2021},
  organization={PmLR}
}

@inproceedings{zhou2022detecting,
  title={Detecting twenty-thousand classes using image-level supervision},
  author={Zhou, Xingyi and Girdhar, Rohit and Joulin, Armand and Kr{\"a}henb{\"u}hl, Philipp and Misra, Ishan},
  booktitle={European conference on computer vision},
  pages={350--368},
  year={2022},
  organization={Springer}
}

@article{liu2022bevfusion,
  title={Bevfusion: Multi-task multi-sensor fusion with unified bird's-eye view representation},
  author={Liu, Zhijian and Tang, Haotian and Amini, Alexander and Yang, Xinyu and Mao, Huizi and Rus, Daniela and Han, Song},
  journal={arXiv preprint arXiv:2205.13542},
  year={2022}
}

@article{minderer2023scaling,
  title={Scaling open-vocabulary object detection},
  author={Minderer, Matthias and Gritsenko, Alexey and Houlsby, Neil},
  journal={Advances in Neural Information Processing Systems},
  volume={36},
  pages={72983--73007},
  year={2023}
}

@article{hu2024metric3d,
  title={Metric3d v2: A versatile monocular geometric foundation model for zero-shot metric depth and surface normal estimation},
  author={Hu, Mu and Yin, Wei and Zhang, Chi and Cai, Zhipeng and Long, Xiaoxiao and Chen, Hao and Wang, Kaixuan and Yu, Gang and Shen, Chunhua and Shen, Shaojie},
  journal={IEEE Transactions on Pattern Analysis and Machine Intelligence},
  year={2024},
  publisher={IEEE}
}

@inproceedings{yang2024improving,
  title={Improving Distant 3D Object Detection Using 2D Box Supervision},
  author={Yang, Zetong and Yu, Zhiding and Choy, Chris and Wang, Renhao and Anandkumar, Anima and Alvarez, Jose M},
  booktitle={Proceedings of the IEEE/CVF Conference on Computer Vision and Pattern Recognition},
  pages={14853--14863},
  year={2024}
}

@inproceedings{yang2023bevformer,
  title={Bevformer v2: Adapting modern image backbones to bird's-eye-view recognition via perspective supervision},
  author={Yang, Chenyu and Chen, Yuntao and Tian, Hao and Tao, Chenxin and Zhu, Xizhou and Zhang, Zhaoxiang and Huang, Gao and Li, Hongyang and Qiao, Yu and Lu, Lewei and others},
  booktitle={Proceedings of the IEEE/CVF conference on computer vision and pattern recognition},
  pages={17830--17839},
  year={2023}
}

@article{li2024bevformer,
  title={Bevformer: learning bird's-eye-view representation from lidar-camera via spatiotemporal transformers},
  author={Li, Zhiqi and Wang, Wenhai and Li, Hongyang and Xie, Enze and Sima, Chonghao and Lu, Tong and Yu, Qiao and Dai, Jifeng},
  journal={IEEE Transactions on Pattern Analysis and Machine Intelligence},
  year={2024},
  publisher={IEEE}
}

@inproceedings{philion2020lift,
  title={Lift, splat, shoot: Encoding images from arbitrary camera rigs by implicitly unprojecting to 3d},
  author={Philion, Jonah and Fidler, Sanja},
  booktitle={European conference on computer vision},
  pages={194--210},
  year={2020},
  organization={Springer}
}

@inproceedings{li2023bevstereo,
  title={Bevstereo: Enhancing depth estimation in multi-view 3d object detection with temporal stereo},
  author={Li, Yinhao and Bao, Han and Ge, Zheng and Yang, Jinrong and Sun, Jianjian and Li, Zeming},
  booktitle={Proceedings of the AAAI Conference on Artificial Intelligence},
  volume={37},
  number={2},
  pages={1486--1494},
  year={2023}
}

@article{park2022time,
  title={Time will tell: New outlooks and a baseline for temporal multi-view 3d object detection},
  author={Park, Jinhyung and Xu, Chenfeng and Yang, Shijia and Keutzer, Kurt and Kitani, Kris and Tomizuka, Masayoshi and Zhan, Wei},
  journal={arXiv preprint arXiv:2210.02443},
  year={2022}
}

@article{xie2023robobev,
  title={Robobev: Towards robust bird's eye view perception under corruptions},
  author={Xie, Shaoyuan and Kong, Lingdong and Zhang, Wenwei and Ren, Jiawei and Pan, Liang and Chen, Kai and Liu, Ziwei},
  journal={arXiv preprint arXiv:2304.06719},
  year={2023}
}

@inproceedings{liu2024ray,
  title={Ray denoising: Depth-aware hard negative sampling for multi-view 3d object detection},
  author={Liu, Feng and Huang, Tengteng and Zhang, Qianjing and Yao, Haotian and Zhang, Chi and Wan, Fang and Ye, Qixiang and Zhou, Yanzhao},
  booktitle={European Conference on Computer Vision},
  pages={200--217},
  year={2024},
  organization={Springer}
}

@inproceedings{zhang2025mr,
  title={Mr. detr: Instructive multi-route training for detection transformers},
  author={Zhang, Chang-Bin and Zhong, Yujie and Han, Kai},
  booktitle={Proceedings of the Computer Vision and Pattern Recognition Conference},
  pages={9933--9943},
  year={2025}
}

@article{lin2024moe,
  title={Moe-llava: Mixture of experts for large vision-language models},
  author={Lin, Bin and Tang, Zhenyu and Ye, Yang and Cui, Jiaxi and Zhu, Bin and Jin, Peng and Huang, Jinfa and Zhang, Junwu and Pang, Yatian and Ning, Munan and others},
  journal={arXiv preprint arXiv:2401.15947},
  year={2024}
}

@INPROCEEDINGS { Argoverse2,
  author = {Benjamin Wilson and William Qi and Tanmay Agarwal and John Lambert and Jagjeet Singh and Siddhesh Khandelwal and Bowen Pan and Ratnesh Kumar and Andrew Hartnett and Jhony Kaesemodel Pontes and Deva Ramanan and Peter Carr and James Hays},
  title = {Argoverse 2: Next Generation Datasets for Self-driving Perception and Forecasting},
  booktitle = {Proceedings of the Neural Information Processing Systems Track on Datasets and Benchmarks (NeurIPS Datasets and Benchmarks 2021)},
  year = {2021}
}

@inproceedings{li2022grounded,
  title={Grounded language-image pre-training},
  author={Li, Liunian Harold and Zhang, Pengchuan and Zhang, Haotian and Yang, Jianwei and Li, Chunyuan and Zhong, Yiwu and Wang, Lijuan and Yuan, Lu and Zhang, Lei and Hwang, Jenq-Neng and others},
  booktitle={Proceedings of the IEEE/CVF conference on computer vision and pattern recognition},
  pages={10965--10975},
  year={2022}
}

@article{estabrooks2004multiple,
  title={A multiple resampling method for learning from imbalanced data sets},
  author={Estabrooks, Andrew and Jo, Taeho and Japkowicz, Nathalie},
  journal={Computational intelligence},
  volume={20},
  number={1},
  pages={18--36},
  year={2004},
  publisher={Wiley Online Library}
}

@article{kang2019decoupling,
  title={Decoupling representation and classifier for long-tailed recognition},
  author={Kang, Bingyi and Xie, Saining and Rohrbach, Marcus and Yan, Zhicheng and Gordo, Albert and Feng, Jiashi and Kalantidis, Yannis},
  journal={arXiv preprint arXiv:1910.09217},
  year={2019}
}

@inproceedings{lin2017focal,
  title={Focal loss for dense object detection},
  author={Lin, Tsung-Yi and Goyal, Priya and Girshick, Ross and He, Kaiming and Doll{\'a}r, Piotr},
  booktitle={Proceedings of the IEEE international conference on computer vision},
  pages={2980--2988},
  year={2017}
}

@inproceedings{cui2019class,
  title={Class-balanced loss based on effective number of samples},
  author={Cui, Yin and Jia, Menglin and Lin, Tsung-Yi and Song, Yang and Belongie, Serge},
  booktitle={Proceedings of the IEEE/CVF conference on computer vision and pattern recognition},
  pages={9268--9277},
  year={2019}
}

@article{ren2020balanced,
  title={Balanced meta-softmax for long-tailed visual recognition},
  author={Ren, Jiawei and Yu, Cunjun and Ma, Xiao and Zhao, Haiyu and Yi, Shuai and others},
  journal={Advances in neural information processing systems},
  volume={33},
  pages={4175--4186},
  year={2020}
}

@inproceedings{lu2025dynamic,
  title={Dynamic-DINO: Fine-Grained Mixture of Experts Tuning for Real-time Open-Vocabulary Object Detection},
  author={Lu, Yehao and Weng, Minghe and Xiao, Zekang and Jiang, Rui and Su, Wei and Zheng, Guangcong and Lu, Ping and Li, Xi},
  booktitle={Proceedings of the IEEE/CVF International Conference on Computer Vision},
  pages={20847--20856},
  year={2025}
}

@inproceedings{liu2024grounding,
  title={Grounding dino: Marrying dino with grounded pre-training for open-set object detection},
  author={Liu, Shilong and Zeng, Zhaoyang and Ren, Tianhe and Li, Feng and Zhang, Hao and Yang, Jie and Jiang, Qing and Li, Chunyuan and Yang, Jianwei and Su, Hang and others},
  booktitle={European conference on computer vision},
  pages={38--55},
  year={2024},
  organization={Springer}
}

@inproceedings{yang2025fomo,
  title={FOMO-3D: Using Vision Foundation Models for Long-Tailed 3D Object Detection},
  author={Yang, Anqi Joyce and Tu, James and Dvornik, Nikita and Li, Enxu and Urtasun, Raquel},
  booktitle={9th Annual Conference on Robot Learning},
  year={2025}
}

@article{ma2023long,
  title={Long-Tailed 3D Detection via Multi-Modal Fusion},
  author={Ma, Yechi and Peri, Neehar and Wei, Shuoquan and Dave, Achal and Hua, Wei and Li, Yanan and Ramanan, Deva and Kong, Shu},
  journal={arXiv preprint arXiv:2312.10986},
  year={2023}
}

@inproceedings{bai2022transfusion,
  title={Transfusion: Robust lidar-camera fusion for 3d object detection with transformers},
  author={Bai, Xuyang and Hu, Zeyu and Zhu, Xinge and Huang, Qingqiu and Chen, Yilun and Fu, Hongbo and Tai, Chiew-Lan},
  booktitle={Proceedings of the IEEE/CVF conference on computer vision and pattern recognition},
  pages={1090--1099},
  year={2022}
}

@inproceedings{yin2021center,
  title={Center-based 3d object detection and tracking},
  author={Yin, Tianwei and Zhou, Xingyi and Krahenbuhl, Philipp},
  booktitle={Proceedings of the IEEE/CVF conference on computer vision and pattern recognition},
  pages={11784--11793},
  year={2021}
}

@article{fang2024eva,
  title={Eva-02: A visual representation for neon genesis},
  author={Fang, Yuxin and Sun, Quan and Wang, Xinggang and Huang, Tiejun and Wang, Xinlong and Cao, Yue},
  journal={Image and Vision Computing},
  volume={149},
  pages={105171},
  year={2024},
  publisher={Elsevier}
}

@inproceedings{wang2023exploring,
  title={Exploring object-centric temporal modeling for efficient multi-view 3d object detection},
  author={Wang, Shihao and Liu, Yingfei and Wang, Tiancai and Li, Ying and Zhang, Xiangyu},
  booktitle={Proceedings of the IEEE/CVF international conference on computer vision},
  pages={3621--3631},
  year={2023}
}

@article{lin2023sparse4d,
  title={Sparse4d v2: Recurrent temporal fusion with sparse model},
  author={Lin, Xuewu and Lin, Tianwei and Pei, Zixiang and Huang, Lichao and Su, Zhizhong},
  journal={arXiv preprint arXiv:2305.14018},
  year={2023}
}

@inproceedings{liu2023petrv2,
  title={Petrv2: A unified framework for 3d perception from multi-camera images},
  author={Liu, Yingfei and Yan, Junjie and Jia, Fan and Li, Shuailin and Gao, Aqi and Wang, Tiancai and Zhang, Xiangyu},
  booktitle={Proceedings of the IEEE/CVF international conference on computer vision},
  pages={3262--3272},
  year={2023}
}

@article{omi2025load,
  title={Load Balancing Mixture of Experts with Similarity Preserving Routers},
  author={Omi, Nabil and Sen, Siddhartha and Farhadi, Ali},
  journal={arXiv preprint arXiv:2506.14038},
  year={2025}
}

@article{bai2023qwen,
  title={Qwen technical report},
  author={Bai, Jinze and Bai, Shuai and Chu, Yunfei and Cui, Zeyu and Dang, Kai and Deng, Xiaodong and Fan, Yang and Ge, Wenbin and Han, Yu and Huang, Fei and others},
  journal={arXiv preprint arXiv:2309.16609},
  year={2023}
}

@inproceedings{he2016deep,
  title={Deep residual learning for image recognition},
  author={He, Kaiming and Zhang, Xiangyu and Ren, Shaoqing and Sun, Jian},
  booktitle={Proceedings of the IEEE conference on computer vision and pattern recognition},
  pages={770--778},
  year={2016}
}

@article{JMLR:v23:21-0524,
  author  = {T. Tony Cai and Rong Ma},
  title   = {Theoretical Foundations of t-SNE for Visualizing High-Dimensional Clustered Data},
  journal = {Journal of Machine Learning Research},
  year    = {2022},
  volume  = {23},
  number  = {301},
  pages   = {1--54},
  url     = {http://jmlr.org/papers/v23/21-0524.html}
}

@inproceedings{zhong2022regionclip,
  title={Regionclip: Region-based language-image pretraining},
  author={Zhong, Yiwu and Yang, Jianwei and Zhang, Pengchuan and Li, Chunyuan and Codella, Noel and Li, Liunian Harold and Zhou, Luowei and Dai, Xiyang and Yuan, Lu and Li, Yin and others},
  booktitle={Proceedings of the IEEE/CVF conference on computer vision and pattern recognition},
  pages={16793--16803},
  year={2022}
}
}


\end{document}